\documentclass[11pt,a4paper]{article}
\usepackage[hyperref]{acl2019}
\usepackage{times}
\usepackage{latexsym}

\usepackage{url}
\aclfinalcopy 
\def\aclpaperid{1760} 

\usepackage{graphicx}
\usepackage{amsmath}
\usepackage{cases}
\usepackage{booktabs}
\usepackage{algorithm}
\usepackage{algorithmic}
\usepackage{subfig}

\usepackage{mystyle}
\usepackage{mlsymbols}

\newcommand{\bv}{\bm{v}}
\newcommand{\ba}{\bm{a}}

\newcommand{\by}{\bm{y}}

\graphicspath{ {images/} }

\definecolor{mdgreen}{rgb}{0.05,0.6,0.05}
\newcommand{\lbl}[1]{\textsc{#1}}
\newcommand{\misc}[1]{\textcolor{mdgreen}{\lbl{#1}}}

\newcommand{\FN}{\misc{Fn}\xspace}
\newcommand{\CHANGE}{\misc{Ct}\xspace}
\newcommand{\SUSTAIN}{\misc{St}\xspace}
\newcommand{\FA}{\misc{Fa}\xspace}
\newcommand{\GI}{\misc{Gi}\xspace}
\newcommand{\RES}{\misc{Res}\xspace}
\newcommand{\REC}{\misc{Rec}\xspace}
\newcommand{\QUC}{\misc{Quc}\xspace}
\newcommand{\QUO}{\misc{Quo}\xspace}
\newcommand{\MIA}{\misc{Mia}\xspace}
\newcommand{\MIN}{\misc{Min}\xspace}

\newcommand{\module}[1]{\textcolor{dkgray}{\lbl{#1}}}
\newcommand{\HGRU}{\module{HGRU}\xspace}

\newcommand{\BiDAFH}{\module{BiDAF$^H$}\xspace}
\newcommand{\GMGRUH}{\module{GMGRU$^H$}\xspace}
\newcommand{\self}{\module{self$_{42}$}\xspace}
\newcommand{\anchor}{\module{anchor$_{42}$}\xspace}

\newcommand{\Paragraph}[1]{\noindent\textbf{#1.}}

\newcommand{\mytitle}{Observing Dialogue in Therapy: \\Categorizing and Forecasting Behavioral Codes}

\title{\mytitle}

\author{
  Jie Cao${^\dagger}$,
  Michael Tanana${^\ddagger}$,
  Zac E. Imel${^\ddagger}$,
  Eric Poitras${^\ddagger}$,\\
  {\bf David C. Atkins}${^\diamondsuit}$,
  {\bf Vivek Srikumar}${^\dagger}$\\
  ${^\dagger}$School of Computing, University of Utah\\
  ${^\ddagger}$Department of Educational Psychology, University of Utah\\
  ${^\diamondsuit}$Department of Psychiatry and Public Health, University of Washington\\
  \texttt{\{jcao, svivek\}@cs.utah.edu,}\\
  \texttt{\{michael.tanana, zac.imel, eric.poitras\}@utah.edu,}\\
  \texttt{datkins@u.washington.edu}
 }

\date{}

\begin{document}
\maketitle

\begin{abstract}
  Automatically analyzing dialogue can help understand and guide
  behavior in domains such as counseling, where interactions are
  largely mediated by conversation. In this paper, we study modeling
  behavioral codes used to asses a psychotherapy treatment style
  called Motivational Interviewing (MI), which is effective for
  addressing substance abuse and related problems. Specifically, we
  address the problem of providing real-time guidance to therapists
  with a dialogue observer that (1) categorizes therapist and client
  MI behavioral codes and, (2) forecasts codes for upcoming utterances
  to help guide the conversation and potentially alert the
  therapist. For both tasks, we define neural network models that
  build upon recent successes in dialogue modeling. Our experiments
  demonstrate that our models can outperform several baselines for
  both tasks.  We also report the results of a careful analysis that
  reveals the impact of the various network design tradeoffs for
  modeling therapy dialogue.
\end{abstract}

\section{Introduction}
\label{sec:intro}
Conversational agents have long been studied in the context of
psychotherapy, going back to chatbots such as
ELIZA~\cite{weizenbaum1966eliza} and
PARRY~\cite{colby1975artificial}. Research in modeling such dialogue
has largely sought to simulate a participant in the conversation.

In this paper, we argue for modeling dialogue \emph{observers}
instead of participants, and focus on psychotherapy. An observer
could help an \emph{ongoing} therapy session in several ways.
First, by monitoring fidelity to therapy standards, a helper could
guide both veteran and novice therapists towards better patient
outcomes. Second, rather than generating therapist utterances, it
could suggest the type of response that is appropriate. Third, it
could alert a therapist about potentially important cues from a
patient.
Such assistance would be especially helpful in the increasingly
prevalent online or text-based counseling services.\footnote{For
  example, Crisis Text Line (\url{https://www.crisistextline.org}),
  7 Cups (\url{https://www.7cups.com}), etc.}

\begin{table*}[ht]
  \begin{center}
\setlength{\tabcolsep}{4pt}
{\small
\begin{tabular}{llll}
  \toprule
  {\bf Code}            & {\bf Count}            & {\bf Description}                                                                                            & {\bf Examples}                                    \\
  \midrule \midrule
  \multicolumn{4}{c}{ \bf Client Behavioral Codes }                                                                                                                                                                 \\
  \midrule
  \multirow{2}{*}{\FN}  & \multirow{2}{*}{47715} & \multirow{2}{*}{\parbox{5.5cm}{Follow/ Neutral: unrelated to changing or sustaining behavior.}}              & ``You know, I didn't smoke for a while.''         \\
                        &                        &                                                                                                              & ``I have smoked for forty years now.''            \\ 
  \CHANGE               & 5099                   & Utterances about changing unhealthy  behavior.                                                                          & ``I want to stop smoking.''                       \\ 
  \SUSTAIN              & 4378                   & Utterances about sustaining unhealthy behavior.                                                                        & ``I really don't think I smoke too much.''        \\ \midrule
  \midrule
  \multicolumn{4}{c}{\bf Therapist Behavioral Codes }                                                                                                                                                               \\
  \midrule
  \FA                   & 17468                  & Facilitate conversation                                                                                      & ``Mm Hmm.'', ``OK.'',``Tell me more.''            \\
  \GI                   & 15271                  & Give information or feedback.                                                                                & ``I'm Steve.'', ``Yes, alcohol is a depressant.'' \\
  \multirow{2}{*}{\RES} & \multirow{2}{*}{6246}  & \multirow{2}{*}{\parbox{5.5cm}{Simple reflection about the client’s most recent utterance.}}                 & C: ``I didn't smoke last week''                   \\
                        &                        &                                                                                                              & T: ``Cool, you avoided smoking last week.''       \\ 
  \multirow{2}{*}{\REC} & \multirow{2}{*}{4651}  & \multirow{2}{*}{\parbox{5.5cm}{Complex reflection based on a client's history or the broader conversation.}} & C: ``I didn't smoke last week.''                  \\
                        &                        &                                                                                                              & T: ``You mean things begin to change''.           \\ 
  \QUC                  & 5218                   & Closed question                                                                                              & ``Did you smoke this week?''                      \\ 
  \QUO                  & 4509                   & Open question                                                                                                & ``Tell me more about your week.''                 \\ 
  \multirow{2}{*}{\MIA} & \multirow{2}{*}{3869}  & \multirow{2}{*}{\parbox{5.5cm}{Other MI adherent,\eg, affirmation, advising with permission, etc.}}          & ``You've accomplished a difficult task.''         \\
                        &                        &                                                                                                              & ``Is it OK if I suggested something?''            \\ 
  \multirow{2}{*}{\MIN} & \multirow{2}{*}{1019}  & \multirow{2}{*}{\parbox{5.5cm}{MI non-adherent, \eg, confrontation, advising without permission, etc.}}      & ``You hurt the baby's health for cigarettes?''    \\
                        &                        &                                                                                                              & ``You ask them not to drink at your house.''      \\\bottomrule
\end{tabular}}
\end{center}
\caption{Distribution, description and examples of MISC labels.} 
\label{tbl:misc}
\end{table*}

We ground our study in a style of therapy called Motivational
Interviewing~\cite[MI,][]{miller2003motivational,miller2012motivational},
which is widely used for treating addiction-related problems.
To help train therapists, and also to monitor therapy quality,
utterances in sessions are annotated using a set of behavioral codes
called Motivational Interviewing Skill
Codes~\cite[MISC,][]{miller2003manual}. Table~\ref{tbl:misc} shows
standard therapist and patient (\ie, client) codes with
examples. Recent NLP work~\cite[][{\em inter
  alia}]{tanana2016comparison, xiao2016behavioral,
  perez2017predicting, huang2018modeling} has studied the problem of
using MISC to assess \emph{completed} sessions.  Despite its
usefulness, automated post hoc MISC labeling does not address the
desiderata for ongoing sessions identified above; such models use
information from utterances yet to be said. To provide real-time
feedback to therapists, we define two complementary dialogue
observers:
\begin{enumerate}[nosep] 
\item \textbf{Categorization}: Monitoring an ongoing session by
  predicting MISC labels for therapist and client utterances as they
  are made.
\item \textbf{Forecasting}: Given a dialogue history, forecasting
  the MISC label for the next utterance, thereby both alerting or
  guiding therapists.
\end{enumerate}
Via these tasks, we envision a helper that offers assistance to a
therapist in the form of MISC labels.

We study modeling challenges associated with these tasks related to:
\begin{inparaenum}[(1)]
\item representing words and utterances in therapy dialogue,
\item ascertaining relevant aspects of utterances and the dialogue
  history, and
\item handling label imbalance (as evidenced in
  Table~\ref{tbl:misc}).
\end{inparaenum}
We develop neural models that address these challenges in this
domain.

Experiments show that our proposed models outperform baselines by a
large margin. For the categorization task, our models even
outperform previous session-informed approaches that use information
from future utterances. For the more difficult forecasting task, we
show that even without having access to an utterance, the dialogue
history provides information about its MISC label.  We also report
the results of an ablation study that shows the impact of the
various design choices.\footnote{The code is available online at
\url{https://github.com/utahnlp/therapist-observer}.}.

In summary, in this paper, we
\begin{inparaenum}[(1)]
\item define the tasks of categorizing and forecasting Motivational
  Interviewing Skill Codes to provide real-time assistance to
  therapists,
\item propose neural models for both tasks that outperform several
  baselines, and
\item show the impact of various modeling choices via extensive
  analysis.
\end{inparaenum}


\section{Background and Motivation} \label{sec:background}
Motivational Interviewing (MI) is a style of psychotherapy that
seeks to resolve a client's ambivalence towards their problems,
thereby motivating behavior change. Several meta-analyses and
empirical studies have shown the high efficacy and success of MI in
psychotherapy~\cite{burke2004emerging, martins2009review,
  lundahl2010meta}. However, MI skills take practice to master and
require ongoing coaching and feedback to
sustain~\cite{Schwalbe2014}.  Given the emphasis on using specific
types of linguistic behaviors in MI (\eg,
open questions and reflections), fine-grained behavioral coding
plays an important role in MI theory and training.

Motivational Interviewing Skill Codes (MISC, table~\ref{tbl:misc})
is a framework for coding MI sessions. It facilitates evaluating
therapy sessions via utterance-level labels that are akin to
dialogue acts~\cite{stolcke2000dialogue,jurafsky2018speech}, and are designed to examine therapist and client
behavior in a therapy session.\footnote{The original MISC description of
  \citet{miller2003manual} included 28 labels (9 client, 19
  therapist). Due to data scarcity and label confusion, various
  strategies are proposed to merge the labels into a coarser set.
  We adopt the grouping proposed by~\citet{xiao2016behavioral}; the
  appendix gives more details.}

As Table~\ref{tbl:misc} shows,
client labels mark utterances as discussing changing or sustaining
problematic behavior (\CHANGE and \SUSTAIN, respectively) or being
neutral (\FN). Therapist utterances are grouped into eight labels,
some of which (\RES, \REC) correlate with improved outcomes, while
MI non-adherent (\MIN) utterances are to be avoided.  MISC labeling
was originally done by trained annotators performing multiple passes
over a session recording or a transcript.  Recent NLP work speeds up
this process by automatically annotating a completed MI
session~\cite[\eg,][]{tanana2016comparison, xiao2016behavioral,
  perez2017predicting}.

\emph{Instead of providing feedback to a therapist after the
  completion of a session, can a dialogue observer provide online
  feedback?} While past work has shown the helpfulness of post hoc
evaluations of a session, prompt feedback would be more helpful,
especially for MI non-adherent responses.  Such feedback opens up
the possibility of the dialogue observer influencing the therapy
session. It could serve as an assistant that offers suggestions to a
therapist (novice or veteran) about how to respond to a client
utterance. Moreover, it could help alert the therapist to
potentially important cues from the client (specifically, \CHANGE or
\SUSTAIN). 

\section{Task Definitions}
\label{sec:task}

In this section, we will formally define the two NLP tasks
corresponding to the vision in \S\ref{sec:background} using the
conversation in table~\ref{tbl:example} as a running example.

Suppose we have an ongoing MI session with utterances
$u_1, u_2,\cdots, u_n$: together, the dialogue history $H_n$.  Each
utterance $u_i$ is associated with its speaker $s_i$, either C
(client) or T (therapist). Each utterance is also associated with
the MISC label $l_i$, which is the object of study. We will refer to
the last utterance $u_n$ as the \emph{anchor}.

We will define two classification tasks over a fixed dialogue
history with $n$ elements --- \emph{categorization} and
\emph{forecasting}. As the conversation progresses, the history will
be updated with a sliding window.  Since the therapist and client
codes share no overlap, we will design separate models for the two
speakers, giving us four settings in all.

\begin{table}[tp]
  \begin{center}
    \setlength{\tabcolsep}{2pt}
    {\small
      \begin{tabular}{crll}
        \toprule
        $i$        & $s_{i}$ & $u_{i}$                             & $l_{i}$  \\ \hline
        1        & T:      & Have you used drugs recently?       & \QUC     \\
        2        & C:      & I stopped for a year, but relapsed. & \FN      \\
        3        & T:      & You will suffer if you keep using. & \MIN     \\
        4        & C:      & Sorry, I just want to quit.         & \CHANGE  \\
        $\cdots$ &         & $\cdots$                            & $\cdots$ \\ \bottomrule
      \end{tabular}
    }
  \end{center}
  \caption{\label{tbl:example} An example of ongoing therapy session}
\end{table}

\Paragraph{Task 1: Categorization}
The goal of this task is to provide real-time feedback to a
therapist during an ongoing MI session. In the running example, the
therapist's confrontational response in the third utterance is not
MI adherent (\MIN); an observer should flag it as such to bring the
therapist back on track. The client's response, however, shows an
inclination to change their behavior (\CHANGE). Alerting a therapist
(especially a novice) can help guide the conversation in a direction
that encourages it.

In essence, we have the following real-time classification task:
\emph{Given the dialogue history $H_n$ which includes the speaker
  information, predict the MISC label $l_n$ for the last utterance
  $u_n$.}

The key difference from previous work in predicting MISC labels is
that we are restricting the input to the
real-time setting. As a result, models can only use the dialogue
history to predict the label, and in particular, we can not use models
such as a conditional random field or a bi-directional LSTM that need both past and future inputs. 


\Paragraph{Task 2: Forecasting}
A real-time therapy observer may be thought of as an expert
therapist who guides a session with suggestions to the therapist.
For example, after a client discloses their recent drug use
relapse, a novice therapist may respond in a confrontational manner
(which is not recommended, and hence coded \MIN). On the other
hand, a seasoned therapist may respond with a complex reflection
(\REC) such as \emph{``Sounds like you really wanted to give up and
  you're unhappy about the relapse.''}
Such an expert may also anticipate important cues
from the client.

The forecasting task seeks to mimic the intent of such a seasoned therapist:
\emph{Given a dialogue history $H_n$ and the next speaker's identity
  $s_{n+1}$, predict the MISC code $l_{n+1}$ of the yet unknown next
  utterance $u_{n+1}$.}

The MISC forecasting task is a previously unstudied problem. 
We argue that forecasting the type of the next utterance, rather than
selecting or generating its text as has been the focus of several
recent lines of work~\cite[\eg,][]{schatzmann2005quantitative,ubuntu,DSTC7},
allows the human in the loop (the therapist) the freedom to
creatively participate in the conversation within the parameters
defined by the seasoned observer, and perhaps even rejecting
suggestions. Such an observer could be especially helpful for
training therapists~\cite{imel2017technology}.
The forecasting task is also related to recent work on detecting
antisocial comments in online
conversations~\cite{zhang2018conversations} whose goal is to
provide an early warning for such events.

\section{Models for MISC Prediction}
\label{sec:devices}



Modeling the two tasks defined in \S\ref{sec:task} requires
addressing four questions:
\begin{inparaenum}[(1)]
\item How do we encode a dialogue and its utterances?
\item Can we discover discriminative words in each utterance?
\item Can we discover which of the previous utterances are relevant?
\item How do we handle label imbalance in our data?
\end{inparaenum}
Many recent advances in 
neural networks can
be seen as plug-and-play components. To facilitate the comparative
study of models, we will describe components that address the
above questions.  
%
In the rest of the paper, we will use \textbf{boldfaced} terms to
denote vectors and matrices and \module{small caps} to denote
component names.

\subsection{Encoding Dialogue}
\label{ssec:dialog_rep}




Since both our tasks are classification tasks over a dialogue
history, our goal is to convert the sequence of utterences into a
single vector that serves as input to the final classifier.

We will use a hierarchical recurrent encoder~\cite[and
others]{li2015hierarchical,sordoni2015hierarchical,serban2016building}
to encode dialogues, specifically a hierarchical gated recurrent
unit (\HGRU) with an utterance and a dialogue encoder. We use a
bidirectional GRU over word embeddings to encode utterances. As is
standard, we represent an utterance $u_i$ by concatenating the final
forward and reverse hidden states. We will refer to this utterance
vector as $\bm{v}_i$. Also, we will use the hidden states of
each word as inputs to the attention components in
\S\ref{ssec:word_att}. We will refer to such contextual word
encoding of the $j^{th}$ word as $\bm{v}_{ij}$. The dialogue encoder
is a unidirectional GRU that operates on a concatenation of
utterance vectors $\bm{v}_i$ and a trainable vector representing the
speaker $s_i$.\footnote{For the dialogue encoder, we use a
  unidirectional GRU because the dialogue is incomplete. For words,
  since the utterances are completed, we can use a BiGRU.} The final
state of the GRU aggregates the entire dialogue history into a
vector $\bm{H}_n$.



The \HGRU skeleton can be optionally augmented with the
word and dialogue attention described next. All the models we will study are two-layer MLPs over the
vector $\bm{H}_n$ that use a ReLU hidden layer and a softmax layer
for the outputs.


\subsection{Word-level Attention}
\label{ssec:word_att}

Certain words in the utterance history are important to categorize or
forecast MISC labels. The identification of these words may depend on
the utterances in the dialogue. For example, to identify that an
utterance is a simple reflection (\RES) we may need to discover that
the therapist is mirroring a recent client utterance; the example
in table~\ref{tbl:misc} illustrates this. Word attention offers a
natural mechanism for discovering such patterns.


We can unify a broad collection of attention mechanisms in NLP under
a single high level architecture~\cite{galassi2019attention}. We
seek to define attention over the word encodings $\bv_{ij}$ in the
history (called queries), guided by the word encodings in the anchor
$\bv_{nk}$ (called keys). The output is a sequence of
attention-weighted vectors, one for each word in the $i^{th}$
utterance.  The $j^{th}$ output vector $\ba_j$ is computed as a
weighted sum of the keys:
\begin{equation}
  \ba_{ij} = \sum_{k} \alpha^{k}_{j} \bv_{nk}
\label{eq:att_sum}
\end{equation}
The weighting factor $\alpha^k_j$ is the attention weight between
the $j^{th}$ query and the $k^{th}$ key, computed as
\begin{equation}
\label{eq:att_weight}
\alpha^{k}_{j} = \frac{\exp\left(f_{m}(\bv_{nk}, \bv_{ij})\right)}{\sum_{j^{\prime}} \exp\left(f_{m}(\bv_{nk}, \bv_{ij^\prime})\right)}
\end{equation}
Here, $f_m$ is a match scoring function between the corresponding
words, and different choices give us different attention mechanisms.

Finally, a combining function $f_{c}$ combines the original word
encoding $\bm{v}_{ij}$ and the above attention-weighted word vector
$\bm{a}_{ij}$ into a new vector representation $\bm{z}_{ij}$ as the final
representation of the query word encoding:
\begin{equation}
\bm{z}_{ij}= f_{c}(\bm{v}_{ij}, \bm{a}_{ij})
\end{equation}

The attention module, identified by the choice of the functions
$f_m$ and $f_c$, converts word encodings in each utterance
$\bv_{ij}$ into attended word encodings $\bm{z}_{ij}$. To use them
in the \HGRU skeleton, we will encode them a second time using a
BiGRU to produce attention-enhanced utterance vectors. For brevity,
we will refer to these vectors as $\bv_i$ for the utterance
$u_i$. If word attention is used, these attended vectors will be
treated as word encodings.

\begin{table}[t]
\begin{center}
  \setlength{\tabcolsep}{3pt}
  {\small
    \begin{tabular}{ll|l}
      \toprule
      Method                 & $f_{m} $                                    & $f_{c}$                                                             \\ \hline
      BiDAF                  & \multirow{2}{*}{$\bm{v}_{nk} {\bm{v}_{ij}^{T}}$}             & $[\bm{v}_{ij};~\bm{a}_{ij};  $                                      \\
                             &                                             & $~~\bm{v}_{ij} \odot \bm{a}_{ij};~\bm{v}_{ij}\odot \bm{a}^{\prime}]$ \\ \hline
      \multirow{2}{*}{GMGRU} & $\bm{w}^{e} \tanh(\bm{W}^{k}\bm{v}_{nk}$    & \multirow{2}{*}{$[\bm{v}_{ij};\bm{a}_{ij}]$}                        \\
                             & $~~+ \bm{W}^{q}[\bm{v}_{ij}; \bm{h}_{j-1}])$ &                                                                     \\ \hline
    \end{tabular}
  }
\end{center}
\caption{\label{tbl:word_att} Summary of word attention mechanisms.
  We simplify BiDAF with multiplicative attention between  word
  pairs for $f_{m}$, while GMGRU uses additive attention
  influenced by the GRU hidden state.
  The vector $\bm{w}_{e} \in\mathbb{R}^{d}$, and matrices
  $\bm{W}^{k}\in \mathbb{R}^{d \times d}$ and
  $\bm{W}^{q} \in\mathbb{R}^{2d \times 2d}$ are parameters of the BiGRU. The vector $\bm{h}_{j-1}$
  is the hidden state from the BiGRU in GMGRU at previous position
  $j-1$.  
  For combination function, BiDAF concatenates bidirectional
  attention information from both the key-aware query vector
  $\ba_{ij}$ and a similarly defined query-aware key vector
  $\ba^{\prime}$. GMGRU uses simple concatenation for $f_c$.}
\end{table}

To complete this discussion, we need to instantiate the two
functions. We use two commonly used attention mechanisms:
BiDAF~\cite{bidaf} and gated matchLSTM~\cite{wang2017gated}. For
simplicity, we replace the sequence encoder in the latter with a
BiGRU and refer to it as GMGRU. Table~\ref{tbl:word_att} shows the
corresponding definitions of $f_{c}$ and $f_{m}$. We refer the
reader to the original papers for further details. In subsequent
sections, we will refer to the two attended versions of the \HGRU as
\BiDAFH and \GMGRUH.

\subsection{Utterance-level Attention}
\label{ssec:sentence_att}
While we assume that the history of utterances is available for both
our tasks, not every utterance is relevant to decide a MISC label.
For categorization, the relevance of an utterance to the anchor may
be important. For example, a complex reflection (\REC) may depend on
the relationship of the current therapist utterance to one or more
of the previous client utterances. For forecasting, since we do not
have an utterance to label, several previous utterances may
be relevant. For example, in the conversation in
Table~\ref{tbl:example}, both $u_2$ and $u_4$ may be used to
forecast a complex reflection.

To model such utterance-level attention, we will employ the
multi-head, multi-hop attention mechanism used in Transformer
networks~\cite{NIPS2017_7181}. As before, due to space constraints,
we refer the reader to the original work for details. We will use
the $(\bm{Q}, \bm{K}, \bm{V})$ notation from the original paper
here. These matrices represent a query, key and value
respectively. The multi-head attention is defined as:
\begin{equation}
\label{eq:multihead_attention}
{\small \text{Multihead}(\bm{Q},\bm{K},\bm{V}) = [\text{head}_{1};\cdots; \text{head}_{h}]\bm{W}^{O}}
\end{equation}
\begin{equation*}
 \text{head}_{i} = \text{softmax}\left(\frac{\bm{Q}\bm{W}^{Q}_{i}\left(\bm{K}\bm{W}^{K}_{i}\right)^T}{\sqrt{d_{k}}}\right)\bm{V}\bm{W}^{V}_{i}
\end{equation*}
The $\bm{W}_i$'s refer to projection matrices for the three inputs,
and the final $\bm{W}^o$ projects the concatenated heads into a
single vector.

The choices of the query, key and value defines the attention
mechanism. In our work, we compare two variants: {\em anchor-based
  attention}, and {\em self-attention}. The anchor-based attention
is defined by $Q = [\bm{v}_{n}]$ and
$K=V=[\bm{v}_{1} \cdots \bm{v}_{n}]$.  Self-attention is defined by
setting all three matrices to $[\bm{v}_{1} \cdots \bm{v}_{n}]$.
For both settings, we use four heads and stacking them for two hops,
and refer to them as $\self$ and $\anchor$.



\subsection{Addressing Label Imbalance}
\label{ssec:focal_loss}
From Table \ref{tbl:misc}, we see that both client and therapist
labels are imbalanced. Moreover, rarer labels are more important in
both tasks. For example, it is important to identify \CHANGE and
\SUSTAIN utterances. For therapists, it is crucial to flag MI
non-adherent (\MIN) utterances; seasoned therapists are trained to
avoid them because they correlate negatively with patient
improvements. If not explicitly addressed, the frequent but less
useful labels can dominate predictions.


To address this, we extend the focal loss~\cite[FL][]{lin2017focal}
to the multiclass case. For a label $l$ with probability produced by
a model $p_t$, the loss is defined as 
\begin{equation}
 \label{eq:focal}
\text{FL}(p_{t}) = -\alpha_{t} {(1 -p_{t})}^{\gamma} \log(p_{t})
\end{equation}
In addition to using a label-specific balance weight $\alpha_t$, the
loss also includes a modulating factor ${(1-p_{t})}^{\gamma}$ to
dynamically downweight well-classified examples with
$p_{t}\gg0.5$. Here, the $\alpha_t$'s and the $\gamma$ are
hyperparameters. We use FL as the default loss function for all our
models.





\section{Experiments}
\label{sec:experiments}
 The original psychotherapy sessions were collected for both clinical trials and Motivational Interviewing dissemination studies including hospital
settings~\cite{roy2014brief}, outpatient
clinics~\cite{baer2009agency}, college alcohol
interventions~\cite{tollison2008questions, neighbors2012randomized,
  lee2013indicated, lee2014randomized}.  All sessions were annotated with the Motivational Interviewing Skills Codes (MISC) \cite{atkins2014scaling}.    We use the train/test split
of \citet{can2015dialog, tanana2016comparison} to give 243 training MI
sessions and 110 testing sessions. We used 24 training sessions for
development.
As mentioned in \S\ref{sec:background}, all our experiments are
based on the MISC codes grouped by \citet{xiao2016behavioral}.

\subsection{Preprocessing and Model Setup}


An MI session contains about 500 utterances on average. We use a
sliding window of size $N=8$ utterances with padding for the initial
ones. We assume that we always know the identity of the speaker for
all utterances. Based on this, we split the sliding windows into a
client and therapist windows to train separate models.
We tokenized and lower-cased utterances using
spaCy~\cite{spacy2}. To embed words, we concatenated 300-dimensional
Glove embeddings~\cite{pennington2014glove} with ELMo
vectors~\cite{Peters:2018}. The appendix details 
the model setup and hyperparameter choices.




\subsection{Results}
\Paragraph{Best Models}
\label{ssec:models}
Our goal is to discover the best client and therapist models for the
two tasks.  We identified the following best configurations using $\text{F}_{1}$
score on the development set:
\begin{enumerate}[nosep]
\item {\bf Categorization}: For client, the best model does not need any
  word or utterance attention. For the therapist, it uses \GMGRUH
  for word attention and \anchor for utterance attention. We refer
  to these models as $\mathcal{C}_C$ and $\mathcal{C}_T$ respectively
\item {\bf Forecasting}: For both client and  therapist, the best
  model uses no word attention, and uses \self utterance
  attention. We refer to these models as $\mathcal{F}_C$ and
  $\mathcal{F}_T$ respectively.
\end{enumerate}
Here, we show the performance of these models against various
baselines. The appendix gives label-wise precision, recall and $\text{F}_{1}$
scores.



\Paragraph{Results on Categorization}
Tables~\ref{tbl:main_rst_c_categorizing} and
\ref{tbl:main_rst_t_categorizing} show the performance of the
$\mathcal{C}_C$ and $\mathcal{C}_T$ models and the baselines.
For both therapist and client categorization, we compare the best
models against the same set of baselines. The majority baseline
illustrates the severity of the label imbalance
problem. \citet{xiao2016behavioral}, $\text{BiGRU}_{\text{generic}}$,
\citet{can2015dialog} and \citet{tanana2016comparison} are the
previous published baselines. The best results of previous published
baselines are underlined. The last row
$\Delta$ in each table lists the changes of our best model from
them. $\text{BiGRU}_{\text{ELMo}}$, $\text{CONCAT}^{C}$,
$\text{GMGRU}^{H}$ and $\text{BiDAF}^{H}$ are new baselines
we define below.

\begin{table}[!h]
\begin{center}{\small
\setlength{\tabcolsep}{3pt}
\begin{tabular}{lcccc}
\toprule
Method                                         & macro                & \FN                    & \CHANGE              & \SUSTAIN             \\ \midrule
Majority                                       & 30.6                 & {\bf \underline{91.7}} & 0.0                  & 0.0                  \\
\citet{xiao2016behavioral}                     & 50.0                 & 87.9                   & 32.8                 & \underline{29.3}     \\
$\text{BiGRU}_{\text{generic}}$                & \underline{50.2}     & 87.0                   & \underline{35.2}     & 28.4                 \\
$\text{BiGRU}_{\text{ELMo}}$                   & 52.9                 & 87.6                   & {\bf 39.2}           & 32.0                 \\
\midrule
\citet{can2015dialog}                          & 44.0                 & 91.0                   & 20.0                 & 21.0                 \\
\citet{tanana2016comparison}                   & 48.3                 & 89.0                   & 29.0                 & 27.0                 \\
$\text{CONCAT}^{C}$                            & 51.8                 & 86.5                   & 38.8                 & 30.2                 \\
$\text{GMGRU}^{H}$                             & 52.6                 & 89.5                   & 37.1                 & 31.1                 \\
$\text{BiDAF}^{H}$                             & 50.4                 & 87.6                   & 36.5                 & 27.1                 \\ \midrule
  $\mathcal{C}_{C}$                            & {\bf 53.9}           & 89.6                   & 39.1                 & {\bf 33.1}           \\
$\Delta=\mathcal{C}_{C} - \text{\underline{score}}$ & {\footnotesize +3.5} & {\footnotesize -2.1}   & {\footnotesize +3.9} & {\footnotesize +3.8} \\
  \bottomrule
\end{tabular}}
\end{center}
\caption{\label{tbl:main_rst_c_categorizing} Main results on categorizing
  client codes, in terms of macro $\text{F}_{1}$, and $\text{F}_{1}$ for
  each client code. Our model $\mathcal{C}_C$ uses final dialogue
  vector $H_{n}$ and current utterance vector $v_{n}$ as input of MLP
  for final prediction. We found that predicting using
  $\text{MLP}(H_{n})+\text{MLP}(v_{n})$ performs better than just
  $\text{MLP}({H_{n}})$.}
\end{table}

\begin{table*}[!htp]
\begin{center}{\small
\begin{tabular}{lccccccccc}
\toprule
Method                                         & macro                & \FA                  & \RES                 & \REC             & \GI              & \QUC             & \QUO             & \MIA             & \MIN             \\ \midrule
Majority                                       & 5.87                 & 47.0                 & 0.0                  & 0.0              & 0.0              & 0.0              & 0.0              & 0.0              & 0.0              \\
\citet{xiao2016behavioral}                     & 59.3                 & \underline{94.7}     & 50.2                 & 48.3             & 71.9             & 68.7             & 80.1             & 54.0             & 6.5              \\
$\text{BiGRU}_{\text{generic}}$                & \underline{60.2}     & 94.5                 & \underline{50.5}     & \underline{49.3} & 72.0             & 70.7             & 80.1             & \underline{54.0} & \underline{10.8} \\
$\text{BiGRU}_{\text{ELMo}}$                   & 62.6                 & 94.5                 & 51.6                 & 49.4             & 70.7             & 72.1             & 80.8             & 57.2             & 24.2             \\ \midrule
\citet{can2015dialog}                          & -                    & 94.0                 & 49.0                 & 45.0             & \underline{74.0} & \underline{72.0} & \underline{81.0} & -                & -                \\
\citet{tanana2016comparison}                   & -                    & 94.0                 & 48.0                 & 39.0             & 69.0             & 68.0             & 77.0             & -                & -                \\
$\text{CONCAT}^{C}$                            & 61.0                 & 94.5                 & 54.6                 & 34.3             & 73.3             & 73.6             & 81.4             & 54.6             & 22.0             \\
$\text{GMGRU}^{H}$                             & 64.9                 & 94.9                 & {\bf 56.0}           & 54.4             & {\bf 75.5}       & {\bf 75.7}       & {\bf 83.0}       & {\bf 58.2}       & 21.8             \\
$\text{BiDAF}^{H}$                             & 63.8                 & 94.7                 & 55.9                 & 49.7             & 75.4             & 73.8             & 80.7             & 56.2             & 24.0             \\ \midrule
$\mathcal{C}_{T}$                              & {\bf 65.4}           & {\bf 95.0}           & 55.7                 & {\bf 54.9}       & 74.2             & 74.8             & 82.6             & 56.6             & {\bf 29.7}       \\
$\Delta=\mathcal{C}_{T} - \text{\underline{score}}$ & {\footnotesize +5.2} & {\footnotesize +0.3} & {\footnotesize +3.9} & {\footnotesize +3.8} & {\footnotesize +0.2} & {\footnotesize +2.8} & {\footnotesize +1.6} & {\footnotesize +2.6} & {\footnotesize +18.9}                                                                                                \\ \bottomrule
\end{tabular}}
\end{center}
\caption{\label{tbl:main_rst_t_categorizing} Main results on
  categorizing therapist codes, in terms of macro $\text{F}_{1}$, and
  $\text{F}_{1}$ for each therapist code. Models are the same as Table
  ~\ref{tbl:main_rst_c_categorizing}, but tuned for therapist
  codes. For the two grouped MISC set \MIA and \MIN, their results are
  not reported in the original work due to different setting.}
\end{table*}

The first set of baselines (above the line) do not encode dialogue
history and use only the current utterance encoded with a BiGRU. The
work of \citet{xiao2016behavioral} falls in this category, and uses a
100-dimensional domain-specific embedding with weighted cross-entropy
loss. Previously, it was the best model in this class. We also
re-implemented this model to use either ELMo or Glove vectors with
focal loss.\footnote{Other related work in no context
  exists~\cite[e.g.,][]{perez2017predicting, gibson2017attention}, but
  they either do not outperform \cite{xiao2016behavioral} or use
  different data.}

The second set of baselines (below the line) are models that use
dialogue context.  Both \citet{can2015dialog} and
\citet{tanana2016comparison} use well-studied linguistic features and
then tagging the current utterance with both past and future utterance
with CRF and MEMM, respectively. To study the usefulness of the
hierarchical encoder, we implemented a model that uses a bidirectional
GRU over a long sequence of flattened utterance. We refer to this as
$\text{CONCAT}^{C}$. This model is representative of the work
of~\citet{huang2018modeling}, but was reimplemented to take advantage
of ELMo.


For categorizing client codes, $\text{BiGRU}_{\text{ELMo}}$ is a
simple but robust baseline model. It outperforms the previous
best no-context model by more than 2 points on macro $\text{F}_{1}$. Using the
dialogue history, the more sophisticated model $\mathcal{C}_{C}$
further gets 1 point improvement. Especially important is its
improvement on the infrequent, yet crucial labels \CHANGE and
\SUSTAIN. It shows a drop in the $\text{F}_{1}$ on the \FN label, which is
essentially considered to be an unimportant, background class from
the point of view of assessing patient progress.
For therapist codes, as the highlighted numbers in Table
\ref{tbl:main_rst_t_categorizing} show, only incorporating GMGRU-based
word-level attention, $\text{GMGRU}^{H}$ has already outperformed many
baselines, our proposed model $\mathcal{F}_{T}$ which uses both
GMGRU-based word-level attention and anchor-based multi-head multihop
sentence-level attention can further achieve the best overall
performance. Also, note that our models outperform approaches
that take advantage of future utterances.

For both client and therapist codes, concatenating dialogue history
with $\text{CONCAT}^{C}$ always performs worse than the hierarchical
method and even the simpler $\text{BiGRU}_{\text{ELMo}}$.

\begin{table*}[!h]
\centering
\makebox[0pt][c]{\parbox{1.0\textwidth}{%
\begin{minipage}[t]{0.37\textwidth}
\centering
\setlength{\tabcolsep}{1.5pt}
{\small
\begin{tabular}{lcccccc}
  \toprule
  \multirow{2}{*}{Method} & \multicolumn{2}{c}{Dev} & \multicolumn{4}{c}{Test}                                       \\ \cmidrule(lr){2-3} \cmidrule(lr){4-7}
                          & \CHANGE                 & \SUSTAIN   & macro      & \FN        & \CHANGE    & \SUSTAIN   \\ \midrule \midrule
$\text{CONCAT}^{F}$       & 20.4                    & 30.2       & 43.6       & 84.4       & 23.0       & {\bf 23.5} \\
  HGRU                    & 19.9                    & 31.2       & {\bf 44.4} & 85.7       & {\bf 24.9} & 22.5       \\
  $\text{GMGRU}^{H}$      & 19.4                    & 30.5       & 44.3       & 87.1       & 23.3       & 22.4       \\ \midrule
  $\mathcal{F}_{C}$       & {\bf 21.1}              & {\bf 31.3} & 44.3       & 85.2       & 24.7       & 22.7       \\
\bottomrule
\end{tabular}}
\caption*{(a) Main results on forecasting client codes, in terms of $\text{F}_{1}$ for \SUSTAIN, \CHANGE on dev set, and macro $\text{F}_{1}$, and $\text{F}_{1}$ for each client code on the test set.}
\end{minipage}
\hfill
\begin{minipage}[t]{0.60\textwidth}
\centering
\setlength{\tabcolsep}{1.5pt}
{\small
\begin{tabular}{ccccccccccc}
\toprule
\multirow{2}{*}{Method} & \multicolumn{1}{c}{Recall} & \multicolumn{9}{c}{$\text{F}_{1}$}                                                                                             \\ \cmidrule(lr){2-2} \cmidrule(lr){3-11}
                        & R@3                        & macro      & \FA        & \RES       & \REC       & \GI        & \QUC       & \QUO       & \MIA       & \MIN       \\ \midrule \midrule
$\text{CONCAT}^{F}$     & 72.5                       & 23.5       & 63.5       & 0.6        & 0.0        & 53.7       & 27.0       & 15.0       & 18.2       & 9.0        \\
HGRU                    & 76.0                       & 28.6       & 71.4       & 12.7       & {\bf 24.9} & 58.3       & 28.8       & 5.9        & {\bf 17.4} & 9.7        \\
$\text{GMGRU}^{H}$      & 76.6                       & 26.6       & {\bf 72.6} & 10.2       & 20.6       & 58.8       & 27.4       & 6.0        & 8.9        & 7.9        \\ \midrule
$\mathcal{F}_{T}$       & {\bf 77.0}                 & {\bf 31.1} & 71.9       & {\bf 19.5} & 24.7       & {\bf 59.2} & {\bf 29.1} & {\bf 16.4} & 15.2       & {\bf 12.8} \\
\bottomrule
\end{tabular}}
\caption*{(b) Main results on forecasting therapist codes, in terms of Recall@3, macro $\text{F}_{1}$, and $\text{F}_{1}$ for each label on test set}
\end{minipage} %
}}
\caption{\label{tbl:main_rst_forecast} Main results on forecasting task}
\end{table*}

\Paragraph{Results on Forecasting} Since the forecasting task is
new, there are no published baselines to compare against. Our
baseline systems essentially differ in their representation of
dialogue history. The model $\text{CONCAT}^{F}$ uses the same
architecture as the model $\text{CONCAT}^{C}$ from the categorizing
task. We also show comparisons to the simple \HGRU model and the
\GMGRUH model that uses a gated matchGRU for word
attention.\footnote{The forecasting task bears similarity to the
  next utterance selection task in dialogue state tracking
  work~\cite{DSTC7}. In preliminary experiments, we found that the
  Dual-Encoder approach used for that task consistently
  underperformed the other baselines described here.}

Tables \ref{tbl:main_rst_forecast} (a,b) show our forecasting results for
client and therapist respectively. For client codes, we also report
the \CHANGE and \SUSTAIN performance on the development set because
of their importance.  For the therapist codes, we also report the
recall@3 to show the performance of a suggestion system that
displayed three labels instead of one.
The results show that even without an utterance, the dialogue
history conveys signal about the next MISC label. Indeed, the
performance for some labels is even better than some categorization
baseline systems. Surprisingly, word attention ($\text{GMGRU}^{H}$)
in Table \ref{tbl:main_rst_forecast} did not help in forecasting
setting, and a model with the \self utterance attention is
sufficient. For the therapist labels, if we always predicted the
three most frequent labels (\FA, \GI, and \RES), the recall@3 is
only 67.7, suggesting that our models are informative if used in
this suggestion-mode.



\section{Analysis and Ablations}
\label{sec:analysis}
This section reports error analysis and an ablation study of our
models on the development set. The appendix shows a comparison of
pretrained domain-specific ELMo/glove with generic ones and the
impact of the focal loss compared to simple or weighted
cross-entropy.

\subsection{Label Confusion and Error Breakdown}
\label{ssec:label_confusion}

\begin{figure}[!h]
\centering
  \includegraphics[width=0.24\textwidth]{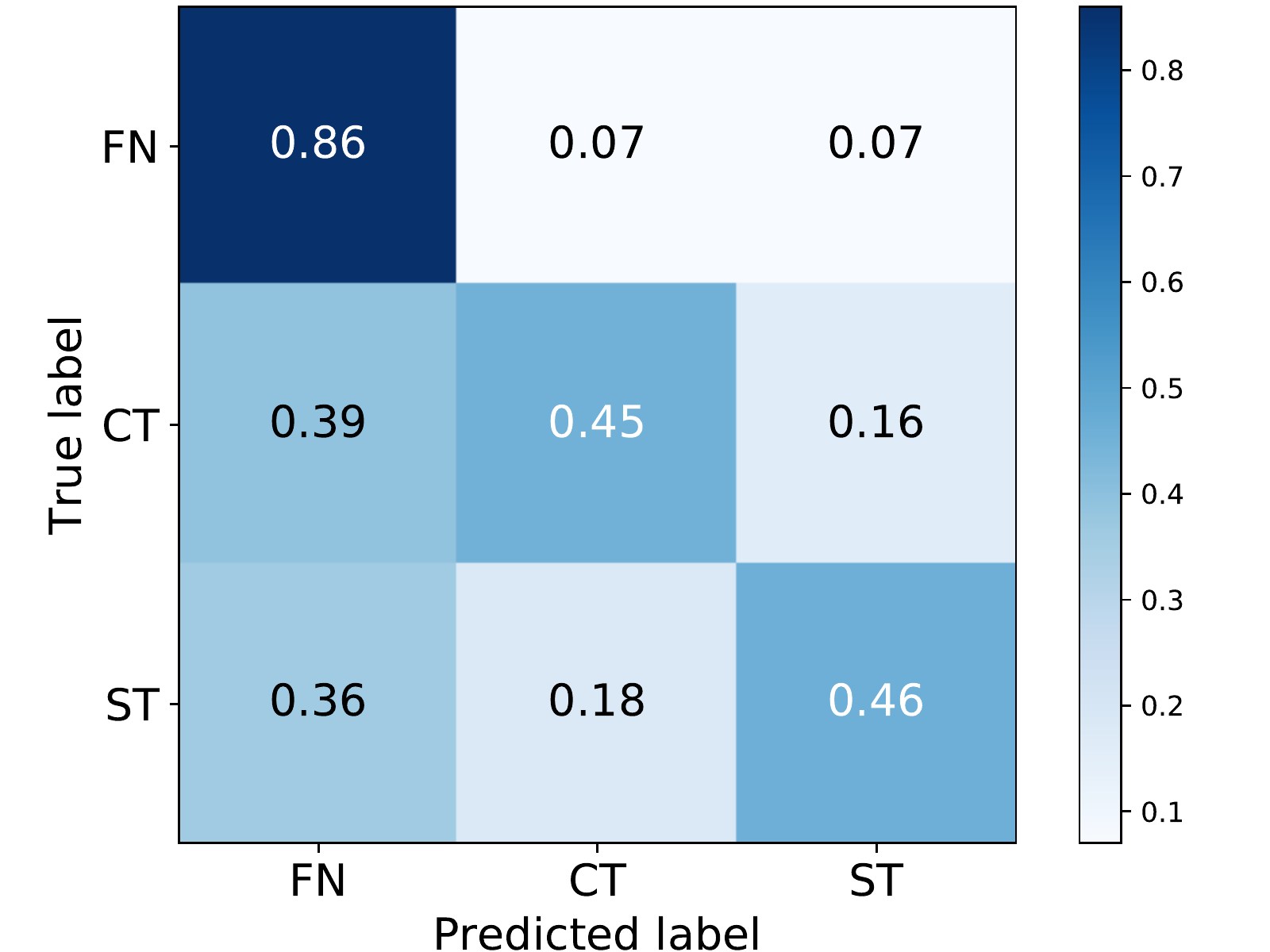}
  \caption{\label{fig:categorizing_confusion_client} Confusion matrix for categorizing client codes, normalized by row.}
\end{figure}

Figure \ref{fig:categorizing_confusion_client} shows the confusion
matrix for the client categorization task. The confusion between \FN
and \CHANGE/\SUSTAIN is largely caused by label imbalance. There are 414 \CHANGE examples that are predicted as \SUSTAIN
and 391 examples vice versa. To further understand their confusion, we
selected 100 of each for manual analysis. We found four broad
categories of confusion, shown in Table
\ref{tbl:c_client_errors}.

\begin{table*}[!h]
  \small
  \begin{center}
\setlength{\tabcolsep}{4pt}
\begin{tabular}{ll}
  \toprule
  {\bf Category and Explaination}                                                                                                                                                                                                                            & {\bf Client Examples (Gold MISC)}                                                                                                                                         \\\midrule
  \multirow{3}{*}{\parbox{7cm}{Reasoning is required to understand
  whether a client wants to change behavior, even with full context~(50,42) }}                               & \multirow{3}{*}{\parbox{8cm}{T: On a scale of zero to ten how confident are you that you can implement this change ? C: I don't know, seven maybe (\CHANGE);\\ I have to wind down after work (\SUSTAIN) }} \\
                                                                                                                                                                                                                                                             &                                                                                                                                                                     \\
                                                                                                                                                                                                                                                             &                                                                                                                                                                     \\\midrule
  \multirow{3}{*}{\parbox{7cm}{Concise utterances which are easy for humans to understand, but missing information such as coreference, zero pronouns~(22,31)}}                                                                                        & I mean I could try it (\CHANGE)                                                                                                                                      \\
                                                                                                                                                                                                                                                             & Not a negative consequence for me (\SUSTAIN)                                                                                                                         \\
                                                                                                                                                                                                                                                             & I want to get every single second and minute out of it(\CHANGE)                                                                                                                         \\\midrule
  \multirow{2}{*}{\parbox{7cm}{Extremely short ($ \leq5 $) or long sentence ($\ge40 $), caused by incorrect turn segementation. ~(21,23)}} & It is a good thing (\SUSTAIN)                                                                                                                                       \\
                                                                                                                                                                                                                                                             & Painful (\CHANGE)                                                                                                                                                    \\ \midrule
  \multirow{2}{*}{\parbox{7cm}{Ambivalent speech, very hard to understand even for human.~(7,4)}}                                                                                                     & What if it does n't work I mean what if I can't do it (\SUSTAIN)                                                                                                                            \\
                                                                                                                                                                                                                                                             & But I can stop whenever I want(\SUSTAIN)                                                                                                                            \\\bottomrule
\end{tabular}
\end{center}
\caption{Categorization of \CHANGE/\SUSTAIN confusions.The two numbers in the brackets are the count of errors for predicting \CHANGE as \SUSTAIN and vice versa. We exampled 100 examples for each case.}
\label{tbl:c_client_errors}
\end{table*}

The first category requires more complex reasoning than just surface
form matching. For example, the phrase \example{seven out of ten}
indicates that the client is very confident about changing behavior;
the phrase \example{wind down after work} indicates, in this context,
that the client drinks or smokes after work. We also found that the
another frequent source of error is incomplete information. In a
face-to-face therapy session, people may use concise and effient
verbal communication, with guestures and other body language conveying
information without explaining details about, for example,
coreference.  With only textual context, it is difficult to infer the
missing information. The third category of errors is introduced when
speech is transcribed into text. The last category is about ambivalent
speech. Discovering the real attitude towards behavior change behind
such utterances could be difficult, even for an expert therapist.
\begin{figure}[!th]
\centering

  \includegraphics[width=0.5\textwidth]{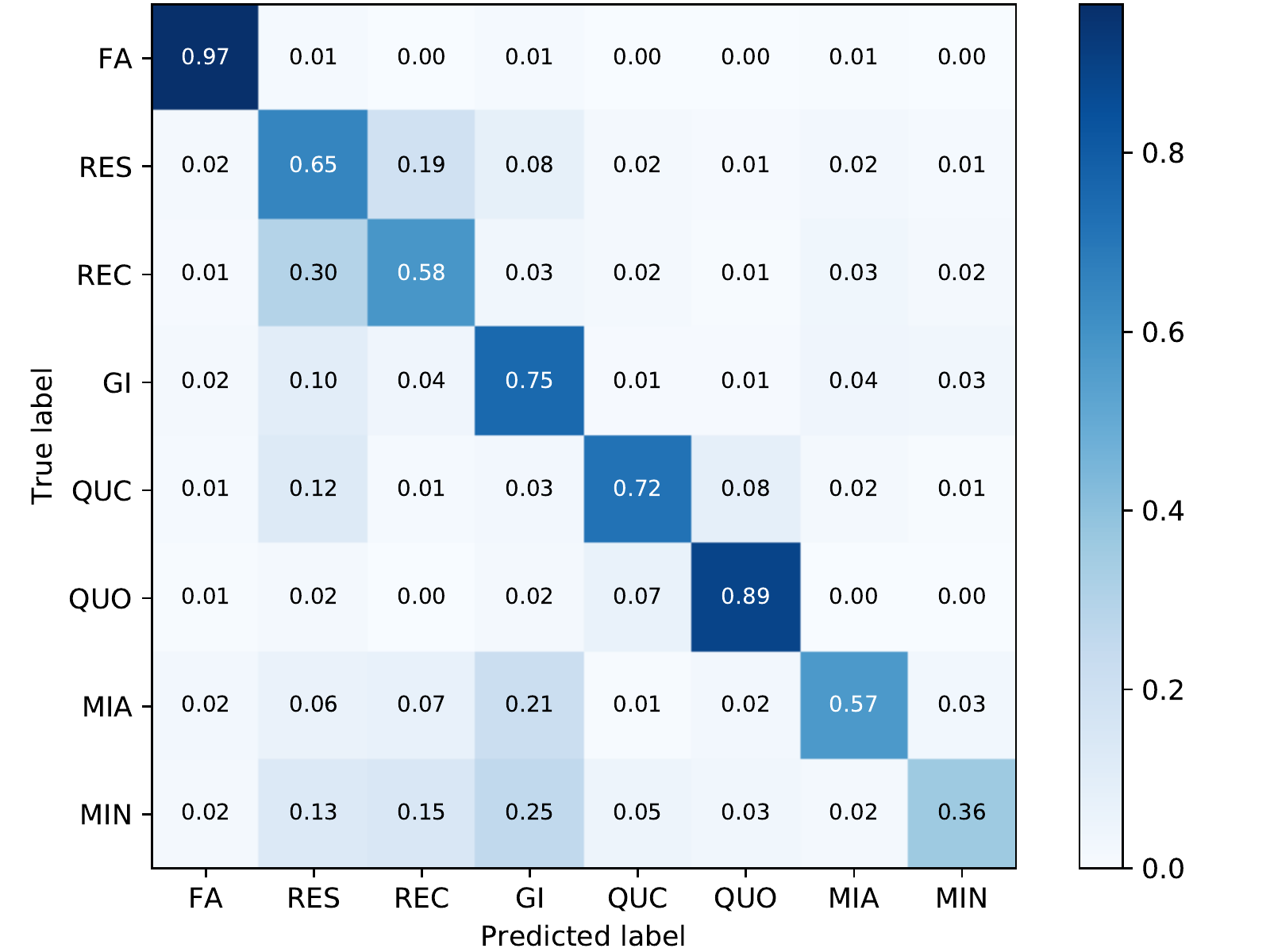}
  \caption{\label{fig:categorizing_confusion_therapist} Confusion matrix for categorizing therapist codes, normalized by row.}
\end{figure}

Figures~\ref{fig:categorizing_confusion_client} and~\ref{fig:categorizing_confusion_therapist} show the
label confusion matrices for the best categorization models. We will
examine confusions that are not caused purely by a label being
frequent. We observe a common confusion between
the two reflection labels, \REC and \RES. Compared to the
confusion matrix from \citet{xiao2016behavioral}, we see that our
models show much-decreased confusion here. There are two reason for
this confusion persisting. First, the reflections may require a much
longer information horizon. We found that by increasing the window
size to 16, the overall reflection results improved. Second, we need
to capture richer meaning beyond surface word overlap for
\RES. We found that
complex reflections usually add meaning or emphasis  to
previous client statements using devices such as analogies,
metaphors, or similes rather than simply restating them. 

%
Closed questions (\QUC) and simple reflections (\RES) are known to
be a confusing set of labels. For example, an utterance like
\example{Sounds like you're suffering?} may be both.
Giving information (\GI)
is easily confused with many labels because they relate to providing
information to clients, but with different attitudes. The MI adherent
(\MIA) and non-adherent (\MIN) labels may also provide information,
but with supportive or critical attitude that may be difficult to
disentangle, given the limited number of examples.



\subsection{How Context and Attention Help?}
\label{ssec:abl_context_attention}

We evaluated various ablations of our best models to see how
changing various design choices changes performance. We focused on
the context window size and impact of different word level and
sentence level attention mechanisms. Tables \ref{tbl:rst_cxt_client}
and \ref{tbl:rst_cxt_therapist} summarize our results.

\Paragraph{History Size}
Increasing the history window size generally helps. The biggest
improvements are for categorizing therapist codes
(Table~\ref{tbl:rst_cxt_therapist}), especially for the \RES and
\REC. However, increasing the window size beyond 8 does not help
to categorize client codes (Table~\ref{tbl:rst_cxt_client}) or
forecasting (in appendix).

\Paragraph{Word-level Attention}
Only the model $\mathcal{C}_{T}$ uses word-level attention. As shown
in Table~\ref{tbl:rst_cxt_therapist}, when we remove the word-level
attention from it, the overall performance drops by 3.4 points,
while performances of \RES and \REC drop by 3.3 and 5 points
respectively. Changing the attention to BiDAF decreases performance
by about 2 points (still higher than the model without attention).

\Paragraph{Sentence-level Attention} Removing sentence attention
from the best models that have it decreases performance for the
models $\mathcal{C}_T$ and $\mathcal{F}_T$ (in appendix).
%
%
  It makes little impact
on the $\mathcal{F}_C$, however.
Table \ref{tbl:rst_cxt_client} shows that neither attention helps
categorizing clients codes.
%

\begin{table}[t]
\begin{center}{\small
\setlength{\tabcolsep}{2pt}
\begin{tabular}{cccccc}
\toprule
 Ablation                                             & Options                      & macro & \FN  & \CHANGE & \SUSTAIN \\ \midrule \midrule
 \multirow{4}{*}{\parbox{1.5cm}{history window size}} & 0                            & 51.6  & 87.6 & 39.2    & 32.0     \\
                                                      & 4                            & 52.6  & 88.5 & 37.8    & 31.5     \\
                                                      & $8^{*}$                      & 53.9  & 89.6 & 39.1    & 33.1     \\
                                                      & 16                           & 52.0  & 89.6 & 39.1    & 33.1     \\ \midrule
\multirow{2}{*}{\parbox{1.5cm}{word \quad \quad attention}}   & + GMGRU                      & 52.6  & 89.5 & 37.1    & 31.1     \\
                                                      & + BiDAF                      & 50.4  & 87.6 & 36.5    & 27.1     \\ \midrule
\multirow{2}{*}{\parbox{1.5cm}{sentence \quad attention}} & + \self                      & 53.9  & 89.2 & 39.1    & 33.2     \\
                                                      & + \anchor                    & 53.0  & 88.2 & 38.9    & 32.0     \\ \bottomrule
\end{tabular}}
\end{center}
\caption{\label{tbl:rst_cxt_client} Ablation study on categorizing
  client code. $*$ is our best model $\mathcal{C}_{C}$. All
  ablation is based on it. The symbol $+$ means adding a
  component to it.  The default window size
  is 8 for our ablation models in the word attention and sentence
  attention parts.}
\end{table}

\begin{table}[t]
\begin{center}{\small
\setlength{\tabcolsep}{3pt}
\begin{tabular}{cccccc}
\toprule
Ablation                                              & Options                        & macro      & \RES       & \REC       & \MIN       \\ \midrule \midrule
 \multirow{4}{*}{\parbox{1.5cm}{history window size}} & 0                              & 62.6       & 51.6       & 49.4       & 24.2       \\
                                                      & 4                              & 64.4       & 54.3       & 53.2       & 23.7       \\
                                                      & $8^{*}$                        & 65.4       & 55.7       & 54.9       & 29.7       \\
                                                      & 16                             & {\bf 65.6} & 55.4       & {\bf 56.7} & 26.7       \\ \midrule
\multirow{2}{*}{\parbox{1.5cm}{word \quad\quad attention}}    & - GMGRU                        & 62.0       & 51.9       & 51.7       & 16.0       \\
                                                      & $\setminus$ BiDAF                      & 63.5       & 54.2       & 51.3       & 22.6       \\\midrule
\multirow{2}{*}{\parbox{1.5cm}{sentence \quad attention}} & - \anchor                      & 64.9       & 56.0       & 54.4       & 21.8       \\
                                                      & $\setminus$ \self                      & 63.4       & 55.5       & 48.2       & 21.1       \\ \bottomrule
\end{tabular}}
\end{center}
\caption{\label{tbl:rst_cxt_therapist} Ablation study on
  categorizing therapist codes, $*$ is our proposed model
  $\mathcal{C}_{T}$. $\setminus$ means substituting and $-$ means removing
  that component. Here, we only report the important \REC, \RES
  labels for
  guiding, and the \MIN label for warning a therapist. }
\end{table}


\subsection{Can We Suggest Empathetic Responses?}

Our forecasting models are trained on regular MI sessions, according
to the label distribution on Table \ref{tbl:misc}, there are both MI
adherent or non-adherent data. Hence, our models are trained to show
how the therapist usually respond to a given statement.

To show whether our model can mimic {\em good} MI policies, we
selected 35 MI sessions from our test set which were rated 5 or
higher on a 7-point scale empathy or spirit. On these sessions, we
still achieve a recall@3 of 76.9, suggesting that we can learn good
MI policies by training on all therapy sessions. These results
suggest that our models can help train new
therapists who may be uncertain about how to respond to a client.






\section{Conclusion}
We addressed the question of providing real-time assistance to
therapists and proposed the tasks of categorizing and forecasting MISC
labels for an ongoing therapy session. By developing a modular family
of neural networks for these tasks, we show that our models outperform several
baselines by a large margin.
Extensive analysis shows
that our model can decrease the label confusion compared to previous
work, especially for reflections and rare labels, but also
highlights directions for future work. 



\section*{Acknowledgments}
The authors wish to thank the anonymous reviewers and members of the
Utah NLP group for their valuable feedback. This research was
supported by an NSF Cyberlearning grant (\#1822877) and a GPU gift
from NVIDIA Corporation.

\bibliography{psyc}
\bibliographystyle{acl_natbib}

\appendix
\section{Appendix}

\paragraph{Different Clustering Strategies for MISC}
\label{ssec:misc_clustering}

\begin{table*}[!h]
  \begin{center}
\setlength{\tabcolsep}{4pt}
{\small
\begin{tabular}{llll}
  \toprule
{\bf Code}           & {\bf Count}            & {\bf Description}                                                                                                                                                                                                     & {\bf Examples}                                      \\ \hline \hline
\multirow{6}{*}{\MIA} & \multirow{6}{*}{3869}  & \multirow{6}{*}{\parbox{5.5cm}{Group of MI Adherent codes : Affirm(\misc{AF}); Reframe(\misc{RF}); Emphasize Control(\misc{EC}); Support(\misc{SU}); Filler(\misc{FI}); Advise with permission(\misc{ADP}); Structure(\misc{ST}); Raise concern with permission(\misc{RCP})}} & ``You've accomplished a difficult task.''~(\misc{\misc{AF}})      \\
                     &                        &                                                                                                                                                                                                                       & ``It’s your decision whether you quit or not''~(\misc{EC}) \\
                     &                        &                                                                                                                                                                                                                       & ``That must have been difficult.''~(\misc{SU})             \\
                     &                        &                                                                                                                                                                                                                       & ``Nice weather today!''~(\misc{FI})                        \\
                     &                        &                                                                                                                                                                                                                       & ``Is it OK if I suggested something?''~(\misc{ADP})        \\
                     &                        &                                                                                                                                                                                                                       & ``Let's go to the next topic''~(\misc{ST})                 \\
                     &                        &                                                                                                                                                                                                                       & ``Frankly, it worries me.''~(\misc{RCP})                   \\  \hline
\multirow{5}{*}{\MIN} & \multirow{5}{*}{1019}  & \multirow{5}{*}{\parbox{5.5cm}{Group of MI Non-adherent codes: Confront(\misc{CO}); Direct(\misc{DI}); Advise without permission(\misc{ADW}); Warn(\misc{WA}); Raise concern without permission(\misc{RCW})}}                                            & ``You hurt the baby's health for cigarettes?''~(\misc{CO}) \\
                     &                        &                                                                                                                                                                                                                       & ``You need to xxx.''~(\misc{DI})                           \\
                     &                        &                                                                                                                                                                                                                       & ``You ask them not to drink at your house.''~(\misc{ADW})  \\
                     &                        &                                                                                                                                                                                                                       & ``You will die if you don't stop smoking.''~(\misc{WA})    \\
                     &                        &                                                                                                                                                                                                                       & ``You may use it again with your friends.''~(\misc{RCW})   \\ \bottomrule
\end{tabular}}
\end{center}
\caption{\label{tbl:misc_mia_min} Label distribution, description and exmaples for \MIA and \MIN}
\end{table*}

The original MISC description of \citet{miller2003manual} included 28
labels (9 client, 19 therapist). Due to data scarcity and label
confusion, some labels were merged into a coarser set.
\citet{can2015dialog} retain 6 original labels \FA, \GI, \QUC, \QUO,
\REC, \RES, and merge remaining 13 rare labels into a single
\misc{COU} label, they merge all 9 client codes into a single
\misc{CLI} label.  Instead, \citet{tanana2016comparison} merge only 8
of rare labels into a \misc{OTHER} label and they cluster client codes
according to the valence of changing, sustaining or being neutral on
the addictive behavior\cite{atkins2014scaling}. Then
\citet{xiao2016behavioral} combine and improve above two clustering
strategies by splitting the all 13 rare labels according to whether
the code represents MI-adherent(\MIA) and MI-nonadherent (\MIN) We
show more details about the original labels in \MIA and \MIN in Table
\ref{tbl:misc_mia_min}

\paragraph{Model Setup}

We use 300-dimensional Glove embeddings pre-trained on 840B tokens
from Common Crawl~\cite{pennington2014glove}. We do not update the
embedding during training. Tokens not covered by Glove are using a
randomly initialized UNK embedding. We also use character-level deep
contextualized embedding ELMo 5.5B model by concatenating the
corresponding ELMo word encoding after the word embedding
vector. For speaker information, we randomly initialize them with 8
dimensional vectors and update them during training. We used a
dropout rate of 0.3 for the embedding layers.

We trained all models using Adam~\cite{kingma2014adam} with learning
rate chosen by cross validation between $[1e^{-4}, 5*1e^{-4}]$,
gradient norms clipping from at $[1.0, 5.0]$, and minibatch sizes of
32 or 64. We use the same hidden size for both utterance encoder,
dialogue encoder and other attention memory hidden size; it has been
selected from $\{64, 128, 256, 512\}$. We set a smaller dropout 0.2
for the final two fully connected layers. All the models are trained
for 100 epochs with early-stoping based on macro $\text{F}_{1}$ over development
results.


\paragraph{Detailed Results of Our Main Models}
\label{ssec:main_rst_prf}
In the main text, we only show the $\text{F}_{1}$ score of each our proposed models.
We summarize the performance of our best models for both categorzing and
forecasting MISC codes in Table~\ref{tbl:main_rst} with precision,
recall and $F_{1}$ for each codes.
\begin{table}[h]
\begin{center}
\setlength{\tabcolsep}{3pt}
\begin{tabular}{l|ccc|ccc}
\toprule
\hline
\multirow{2}{*}{{\bf Label}} & \multicolumn{3}{c}{{\bf Categorizing}} & \multicolumn{3}{c}{{\bf Forecasting}}  \\ \cline{2-4} \cline{5-7}
                             & P                                      & R    & $\text{F}_{1}$ & P    & R    & $\text{F}_{1}$ \\ \hline
\FN                          & 92.5                                   & 86.8 & 89.6    & 90.8 & 80.3 & 85.2    \\
\CHANGE                      & 34.8                                   & 44.7 & 39.1    & 18.9 & 28.6 & 22.7    \\
\SUSTAIN                     & 28.2                                   & 39.9 & 33.1    & 19.5 & 33.7 & 24.7    \\
\hline
\FA                          & 95.1                                   & 94.7 & 94.9    & 70.7 & 73.2 & 71.9    \\
\RES                         & 50.3                                   & 61.3 & 55.2    & 20.1 & 18.8 & 19.5    \\
\REC                         & 52.8                                   & 55.5 & 54.1    & 19.2 & 34.7 & 24.7    \\
\GI                          & 74.6                                   & 75.1 & 74.8    & 52.8 & 67.5 & 59.2    \\
\QUC                         & 80.6                                   & 70.4 & 75.1    & 36.2 & 24.3 & 29.1    \\
\QUO                         & 85.3                                   & 81.2 & 83.2    & 27.0 & 11.8 & 16.4    \\
\MIA                         & 61.8                                   & 52.4 & 56.7    & 27.0 & 10.6 & 15.2    \\
\MIN                         & 27.7                                   & 28.5 & 28.1    & 17.2 & 10.2 & 12.8    \\ \hline \bottomrule
\end{tabular}
\end{center}
\caption{Performance of our proposed
  models with respect to precision, recall and $\text{F}_{1}$ on categorizing and forecasting
  tasks for client and therapist codes}
\label{tbl:main_rst}
\end{table}


\paragraph{Domain Specific Glove and ELMo}
\label{ssec:domain_elmo}
We use the general psychotherapy corpus with 6.5M words (Alexander
Street Press) to train the domain specific  word embeddings
$\textbf{Glove}_{psyc}$ with 50, 100, 300 dimension. Also, we
trained ELMo with 1 highway connection and 256-dimensional output size to get $\textbf{ELMo}_{psyc}$. We found that ELMo 5.5B performs better than ELMo psyc in our experiments, and general Glove-300 is better than the $\textbf{Glove}_{psyc}$. Hence for main results of our models, we use $\textbf{ELMo}_{generic}$ by default.
Please see more details in Table \ref{tbl:rst_elmo}
\begin{table*}[!h]
\begin{center}
\setlength{\tabcolsep}{3pt}
{\small
\begin{tabular}{c|l|cccc|ccccccccc}
  \toprule \hline
Model                          & Embedding                    & macro      & \FN        & \CHANGE    & \SUSTAIN   & macro      & \FA        & \RES       & \REC       & \GI        & \QUC       & \QUO       & \MIA       & \MIN       \\ \hline
\multirow{4}{*}{$\mathcal{C}$} & $\text{ELMo}$                & 53.9       & 89.6       & {\bf 39.1} & {\bf 33.1} & {\bf 65.4} & {\bf 95.0} & {\bf 55.7} & {\bf 54.9} & {\bf 74.2} & {\bf 74.8} & {\bf 82.6} & {\bf 56.6} & {\bf 29.7} \\
                               & $\text{ELMo}_{\text{psyc}}$  & 46.9       & 88.9       & 27.5       & 24.3       & 64.2       & 94.9       & 53.3       & 53.3       & 75.8       & 74.8       & 82.2       & 56.1       & 23.5       \\
                               & $\text{Glove}$               & 50.6       & {\bf 89.9} & 33.4       & 28.6       & 62.2       & 94.6       & 53.7       & 54.2       & 70.3       & 70.0       & 79.1       & 54.7       & 20.9       \\
                               & $\text{Glove}^{\text{pysc}}$ & 47.4       & 88.4       & 23.9       & 30.0       & 63.4       & 94.9       & 54.7       & 52.8       & 75.2       & 71.4       & 80.8       & 53.6       & 23.5       \\ \hline
\multirow{4}{*}{$\mathcal{F}$} & $\text{ELMo}$                & {\bf 44.3} & {\bf 85.2} & {\bf 24.7} & 22.7       & {\bf 31.1} & 71.9       & 19.5       & {\bf 24.7} & {\bf 59.2} & 28.3       & {\bf 17.7} & 15.9       & 9.0        \\
                               & $\text{ELMo}_{\text{psyc}}$  & 43.8       & 84.0       & 22.4       & 25.0       & 29.1       & {\bf 73.5} & 15.5       & 24.3       & 59.1       & {\bf 29.1} & 9.5        & 12.1       & 10.1       \\
                               & $\text{Glove}$               & 42.7       & 83.9       & 21.0       & 23.1       & 30.0       & 72.8       & {\bf 20.8} & 23.7       & 58.2       & 26.2       & 14.5       & 14.5       & 9.6        \\
                               & $\text{Glove}^{\text{pysc}}$ & 43.6       & 81.9       & 23.3       & {\bf 25.7} & 30.8       & 72.1       & 19.7       & 24.4       & 57.3       & 28.9       & 13.7       & {\bf 17.8} & {\bf 23.5} \\
\hline \bottomrule
\end{tabular}}
\end{center}
\caption{\label{tbl:rst_elmo} Ablation study for our proposed model
  with embeddings trained on the psychotherapy corpus.}
\end{table*}


\paragraph{Full Results for Ablation on Forecasting Tasks}
\label{ssec:abl_forecast}

\begin{table*}[t]
\begin{center}
\setlength{\tabcolsep}{2.5pt}
{\small
\begin{tabular}{ccccccccccccc}
\toprule
Ablation                                                               & Options               & \CHANGE    & \SUSTAIN   & R@3        & \FA        & \RES       & \REC       & \GI        & \QUC       & \QUO       & \MIA       & \MIN       \\ \midrule \midrule
\multirow{4}{*}{\parbox{1.5cm}{history size}}                          & 1                     & 17.2       & 15.1       & 66.4       & 59.4       & 12.6       & 9.0        & 44.6       & 16.3       & 14.8       & 11.9       & 4.1        \\
                                                                       & 4                     & 16.8       & 22.6       & 75.3       & 71.4       & 15.6       & 21.1       & 57.1       & {\bf 29.3} & 11.0       & 11.2       & 14.4       \\
                                                                       & $8^{*}$               & 24.7       & 22.7       & {\bf 77.0} & {\bf 72.8} & {\bf 20.8} & 23.1       & 58.1       & 28.3       & {\bf 17.7} & 15.9       & 9.0        \\
                                                                       & 16                    & 23.9       & 20.7       & 76.5       & 71.2       & 13.7       & 24.1       & {\bf 58.5} & 25.9       & 9.7        & 16.2       & 12.7       \\ \midrule
\multirow{2}{*}{\parbox{1.5cm}{\parbox{1.5cm}{word \quad\quad attention}}}     & GMGRU                 & 14.0       & {\bf 23.2} & 75.7       & 71.7       & 14.2       & 23.0       & 57.5       & 26.5       & 8.0        & 15.4       & 11.6       \\
                                                                       & $\text{GMGRU}_{4h}$   & 19.1       & 22.9       & 76.3       & 71.3       & 12.1       & 23.3       & 58.1       & 24.5       & 12.6       & 11.7       & 14.0       \\ \midrule
\multirow{3}{*}{\parbox{1.5cm}{\parbox{1.5cm}{sentence \quad\quad attention}}} & $-$ \self             & {\bf 24.9} & 22.5       & 76.0       & 71.4       & 12.7       & 24.9       & 58.3       & 28.8       & 5.9        & {\bf 17.4} & 9.7        \\
                                                                       & $\setminus$ \anchor           & 22.9       & 22.9       & 76.2       & 72.2       & 15.5       & {\bf 24.6} & 59.5       & 27.1       & 7.7        & 16.3       & 8.3        \\
                                                                       & $+$ GMGRU $\setminus$ \anchor & 6.8        & 23.4       & 76.9       & 70.8       & 8.0        & 24.5       & 58.3       & 24.6       & 10.6       & 14.9       & {\bf 12.1} \\ \midrule
\end{tabular}}
\end{center}
\caption{\label{tbl:rst_cxt_anticipate} Ablation on forecasting task on both client and therapist code. $*$ row are results of our best forecasting model $\mathcal{F}_{C}$, and $\mathcal{F}_{T}$. $\setminus$ means substitute anchor attention with self attention. $+\text{GMGRU}$ \anchor means using word-level attention and achor-based sentence-level attention together. }
\end{table*}

In addition to the ablation table in the main paper for categorizing tasks, we reported more ablation details on forecasting task in Table~\ref{tbl:rst_cxt_anticipate}. Word-level attention shows no help for both client and therapist codes. While sentence-level attention helps more on therapist codes than on client codes. Multi-head self attention alsoachieves better performance than anchor-based attention in forecasting tasks.

\paragraph{Label Imbalance}
\label{ssec:label_imb}
We always use the same $\alpha$ for all weighted focal loss. Besides
considering the label frequency, we also consider the performance
gap between previous reported $\text{F}_{1}$. We
choose to balance weights $\alpha$ as \{1.0,1.0,0.25\} for \CHANGE,\SUSTAIN
and \FN respectively, and \{0.5, 1.0, 1.0, 1.0, 0.75, 0.75,1.0,1.0\}
for \FA, \RES, \REC, \GI, \QUC, \QUO, \MIA, \MIN. As shown in
Table~\ref{tbl:loss}, we report our ablation studies on cross-entropy
loss, weighted cross-entropy loss, and focal loss. Besides the fixed
weights, focal loss offers flexible hyperparameters to weight
examples in different tasks. Experiments shows
that except for the model $\mathcal{C}^{T}$, focal loss outperforms
cross-entropy loss and weighted cross entropy.
\begin{table}[!h]
\setlength{\tabcolsep}{3pt}
\begin{center}{\small
\begin{tabular}{c|ccc|ccccc}
\toprule \hline
\multirow{2}{*}{{\bf Loss}} & \multicolumn{3}{c}{ {\bf Client} } & \multicolumn{5}{c}{ {\bf Therapist} }                    \\\cline{2-4}  \cline{5-9}
                            & $\text{F}_{1}$                            & \CHANGE & \SUSTAIN & $\text{F}_{1}$ & \RES & \REC & \MIA & \MIN \\ \hline \hline
$\mathcal{C}^{{\text{ce}}}$ & 47.0                               & 28.4    & 22.0     & 60.9    & 54.3 & 53.8 & 53.7 & 4.8  \\
$\mathcal{C}^{\text{wce}}$  & 53.5                               & 39.2    & 32.0     & 65.4    & 55.7 & 54.9 & 56.6 & 29.7 \\
$\mathcal{C}^{\text{fl}}$   & 53.9                               & 39.1    & 33.1     & 65.4    & 55.7 & 54.9 & 56.6 & 29.7 \\ \hline
$\mathcal{F}^{{\text{ce}}}$ & 42.1                               & 17.7    & 18.5     & 26.8    & 3.3  & 20.8 & 16.3 & 8.3  \\
$\mathcal{F}^{\text{wce}}$  & 43.1                               & 20.6    & 23.3     & 30.7    & 17.9 & 25.0 & 17.7 & 10.9 \\
$\mathcal{F}^{\text{fl}}$   & 44.2                               & 24.7    & 22.7     & 31.1    & 19.5 & 24.7 & 15.2 & 12.8 \\ \hline
\bottomrule
\end{tabular}}
\end{center}
\caption{\label{tbl:loss} Abalation study of different loss function
  on categorizing and forecasting task. Based on our proposed model for
  our four settings, we compared our best model with crossentropy
  loss(ce), $\alpha$ balanced cross-entropy(wce) and focal loss. Here we
  only report the macro $\text{F}_{1}$ for rare labels and the overall macro
  $\text{F}_{1}$. $\gamma=1$ is the best for both the model $\mathcal{C}_{C}$ and
  $\mathcal{F}_{C}$, while $\gamma=0$ is the best for
  $\mathcal{C}_{T}$ and $\gamma=3$ for $\mathcal{F}_{T}$. Worth to mention,
  when $\gamma=0$, the focal loss degraded into $\alpha$-balanced crossentropy,
  that first two rows are the same for therspit model.}
\end{table}



\end{document}